\documentclass[runningheads,a4paper]{llncs}

\usepackage{amssymb}
\usepackage{graphicx}
\usepackage{times}
\usepackage{epsfig}
\usepackage{amsmath}
\usepackage{amssymb}
\usepackage{cite}
\usepackage{amsmath,amssymb}             
\usepackage{amsfonts}
\usepackage[table,xcdraw]{xcolor}
\usepackage{tabularx}

\usepackage{url}

\newcommand{\argmin}{\operatornamewithlimits{argmin}}

\newcommand\blfootnote[1]{%
  \begingroup
  \renewcommand\thefootnote{}\footnote{#1}%
  \addtocounter{footnote}{-1}%
  \endgroup
}

\begin{document}

\mainmatter  

\title{Rigid Slice-To-Volume Medical Image Registration through Markov Random Fields}

\titlerunning{Rigid Slice-To-Volume Medical Image Registration through Markov Random Fields}

%
%
\author{Roque Porchetto$^1$, Franco Stramana$^1$, Nikos Paragios$^2$, Enzo Ferrante$^2$}
\authorrunning{Roque Porchetto, Franco Stramana, Nikos Paragios, Enzo Ferrante}

\institute{$^1$ UNICEN University, Tandil Argentina\\
$^2$CVN, CentraleSupelec-INRIA, Universite Paris-Saclay, France}


%
%

\toctitle{Rigid Slice-To-Volume Medical Image Registration through Markov Random Fields}
\tocauthor{************}
\maketitle

\begin{abstract}
Rigid slice-to-volume registration is a challenging task, which finds application in medical imaging problems like image fusion for image guided surgeries and motion correction for volume reconstruction. It is usually formulated as an optimization problem and solved using standard continuous methods. In this paper, we discuss how this task be formulated as a discrete labeling problem on a graph. Inspired by previous works on discrete estimation of linear transformations using Markov Random Fields (MRFs), we model it using a pairwise MRF, where the nodes are associated to the rigid parameters, and the edges encode the relation between the variables. We compare the performance of the proposed method to a continuous formulation optimized using simplex, and we discuss how it can be used to further improve the accuracy of our approach. Promising results are obtained using a monomodal dataset composed of magnetic resonance images (MRI) of a beating heart. \blfootnote{Contact: ferrante.enzo@gmail.com}
\end{abstract}

\section{Introduction}
\label{sec:introduction}

Slice-to-volume registration has received increasing attention during the last decades within the medical imaging community. Given a tomographic 2D slice and a 3D volume as input, this challenging problem consists in finding the slice (extracted from the input volume and specified by an arbitrary rigid transformation) that best matches the 2D input image. We stress the fact that we are working with 2D slices (e.g. ultrasound (US)) as opposed to projective 2D images (e.g. X-ray images). This is important since both problems are usually refereed as 2D/3D registration, even if they are intrinsically different. In slice-to-volume registration, every pixel from the 2D image corresponds to a single voxel in 3D space. However, in a projective 2D image every pixel represents a projection of voxels from a given viewpoint.

One can formulate different versions of slice-to-volume registration, depending on several aspects of the problem such as the \textit{matching criterion} used to determine the similarity between the images, the \textit{transformation model} we aim at estimating, the \textit{optimization strategy} used to infer the optimal transformation model (continuous or discrete) and the \textit{number of slices} given as input. In this work, we propose an iconic method (where the matching criterion is defined as a function of the image intensity values) to infer rigid transformation models (specified using 6-DOF). The input consists of a single slice and a single volume, and we formulate it as a discrete optimization problem. 

Discrete methods, where the tasks are usually formulated as a discrete labeling problem on a graph, have become a popular strategy to model vision problems \cite{Wang2013} (and particularly, biomedical vision problems \cite{Paragios2016}) thanks to their modularity, efficiency, robustness and theoretical simplicity. This paper presents a graph-based formulation (inspired by the work of \cite{Zikic2010a, Zikic2010b}) to solve rigid (only) slice-to-volume registration using discrete methods. As we will discuss in section \ref{sec:previousWork}, other works have tackled similar problems. However, to date, no work has shown the potential of discrete methods to deal with rigid slice-to-volume registration. Our main contribution is to put a new spin on graph-based registration theory, by demonstrating that discrete methods and graphical models are suitable to estimate rigid transformations mapping slice-to-volume. We validate our approach using a dataset of magnetic resonance images (MRI) of the heart, and we compare its performance with a state-of-the-art approach based on continuous optimization using simplex method. Moreover, in the spirit of encouraging reproducible research, we make the source code of the application publicly available at the following website: \texttt{
https://gitlab.com/franco.stramana1/slice-to-volume}.

\subsection{Motivation}
\label{sec:motivation}
In the extensive literature of medical image registration, it is possible to identify two main problems which motivated the development of slice-to-volume registration methods during the last decades. The first one is the fusion of pre-operative high-definition volumetric images with intra-operative tomographic slices to perform diagnostic and minimally invasive surgeries. In this case, slice-to-volume registration is one of the enabling technologies for computer-aided image guidance, bringing high-resolution pre-operative data into the operating room to provide more realistic information about the patient's anatomy \cite{Liao2013}. This technique has been used when dealing with several scenarios such as liver surgery \cite{Bao2005}, radio-frequency thermal ablation of prostate cancer \cite{Fei2003}, minimally invasive cardiac procedures \cite{Huang2009}, among many others. 

The second problem is the correction of slicewise motion in volumetric acquisitions. In a variety of situations, inter slice motion may appear when capturing a volumetric image. For example, in case of fetal brain imaging (essential to understand neurodevelopmental disabilities in childhood and infancy) \cite{Rousseau2005}, fetus motion generates inconsistencies due to the slice acquisition time. Another case is related to functional magnetic resonance images (fMRI), usually acquired as time series of multislice single-shot echoplanar images (EPI). Patient head motion during the experiments may introduce artifacts on activation signal analysis. Slice-to-volume registration can be used to alleviate this problem by registering every slice to an anatomical volumetric reference following the well-know map-slice-to-volume (MSV) method \cite{Kim1999}.

\subsection{Previous Work}
\label{sec:previousWork}
Graph-based image registration, where the task is conceived as a discrete labeling problem on a graph, constitutes one of the most efficient and accurate state-of-the-art methods for image registration \cite{Sotiras2013}. Even if they have shown to be particularly suitable to estimate deformable non-linear transformations \cite{Glocker2011, Heinrich2013a}, they were also adapted to the linear case \cite{Zikic2010a}. Most of the publications on the field focus on registering images which are in dimensional correspondence (2D/2D or 3D/3D). In case of projective 2D/3D image registration, only linear transformations were estimated using discrete methods by \cite{Zikic2010b, Zikic2010a}. More recently, several graph-based approaches to perform deformable slice-to-volume registration were introduced in \cite{Ferrante2013, Ferrante2015, Ferrante2015a}. In these works, rigid transformations were computed as a by-product of the deformable parameters, leading to unnecessary computational burden (since rigid transformations are by far lower dimensional than deformable ones). To the best of our knowledge, rigid (only) slice-to-volume registration has not been formulated within this powerful framework. To date, all the methods focusing on this challenging problem are based on continuous (e.g. simplex~\cite{Fei2003}, gradient descent~\cite{Rousseau2005}, Powell's method~\cite{Gholipour2010}, etc.) or heuristic approaches (evolutionary algorithms~\cite{Tadayyon2011}, simulated annealing~\cite{Birkfellner2007}), missing the aforementioned advantages offered by discrete optimization. Based on the work of \cite{Zikic2010b}, we propose a discrete Markov Random Field (MRF) formulation of this problem, delivering more precise results than the state-of-the-art continuous approaches.
Moreover, inspired by the work of \cite{Lempitsky2008} in the context of vector flow estimation, we discuss how continuous state-of-the-art approaches can be used to further refine the rigid transformations obtained through discrete optimization, resulting in more accurate solutions.

\section{Rigid Slice-to-Volume Registration through Markov Random Fields}
\begin{figure}[t]
	\centering
		\includegraphics[width=\textwidth]{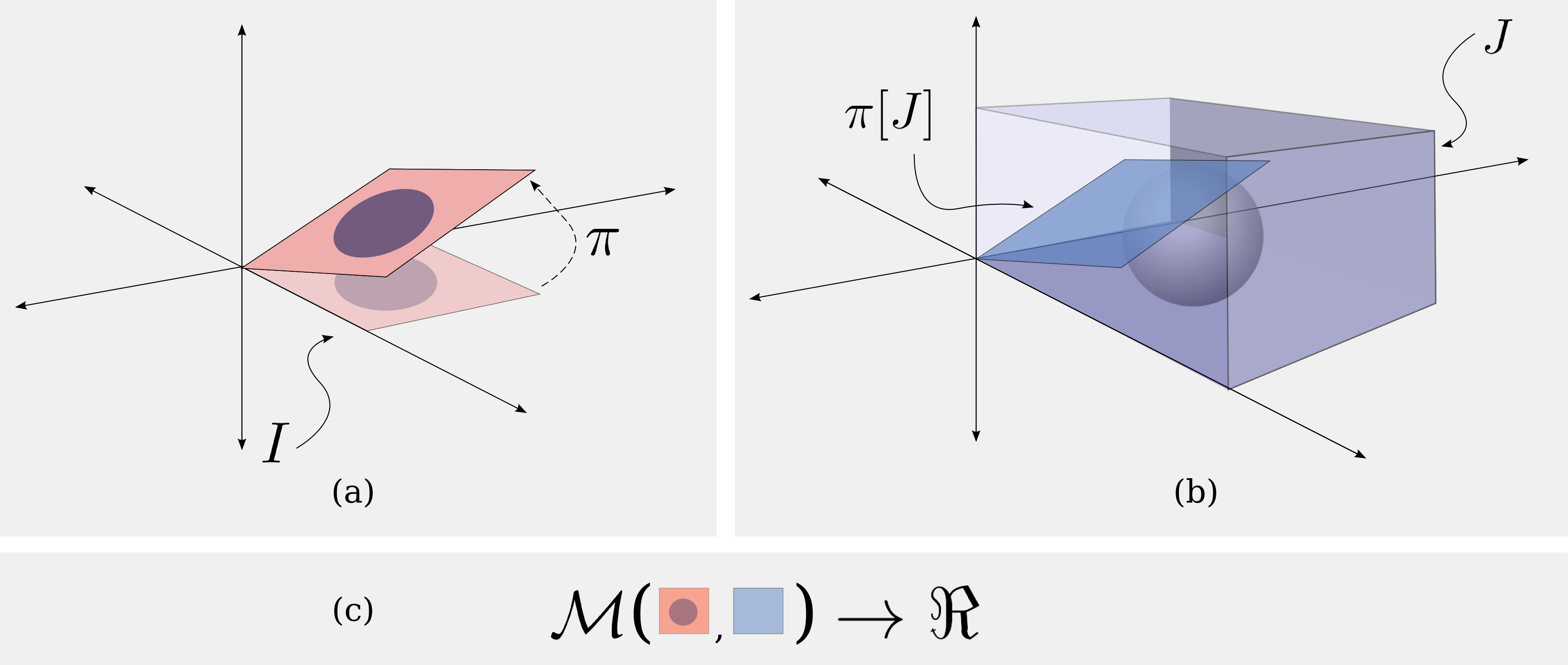}
	\caption{Interpretation of the components of Eq \ref{eq:sliceToVolume}. (a) Image $I$ corresponds to the input 2D image, which is moved by different rigid transformations $\pi$. (b) Image $J$ corresponds to the 3D image. Given a rigid transformation $\pi$, a slice $\pi[J]$ is extracted (using trilinear interpolation). In that way, it is possible to explore the space of solutions by sampling several rigid transformations $\pi$. (c) The matching criterion $\mathcal{M}$ quantifies the dissimilarity of both 2D images, $I$ and $\pi[J]$. Higher values indicate dissimilar images while lower values indicate better alignment.}
	\label{fig:interpretationTransformation}
\end{figure}
We formulate rigid slice-to-volume registration as an optimization problem. Given a 2D image $I: \Omega_{I} \in \Re^2 \rightarrow \Re$, and a 3D image $J: \Omega_J \in \Re^3 \rightarrow \Re$, we aim at recovering the rigid transformation specified by $\pi = (r_x, r_y, r_z, t_x, t_y, t_z)$ that better aligns both images, by solving:

\begin{equation}
	\hat{\pi} = \argmin_{\pi} \mathcal{M}(I, \pi[J]),
	\label{eq:sliceToVolume}
\end{equation}	
where $\pi[J]$ corresponds to the slice extracted from image $J$ (using trilinear interpolation) and specified by the rigid transformation $\pi$ (as explained in Figure \ref{fig:interpretationTransformation}). $\mathcal{M}$ is the so-called matching criteria, that quantifies the dissimilarity between the 2D image $I$ and the slice $\pi[J]$. Alternative matching criteria can be adopted depending on the type of images we are registering. For example, in monomodal cases where intensities tend to be linearly correlated in both images, simple functions such as sum of absolute differences (SAD) or sum of squared differences (SSD) may make the job. However, for more complicated cases like multimodal registration (where the relation between intensity values in both images is usually non-linear), more elaborated functions like mutual information (MI) are applied. 

This optimization problem is commonly solved through continuous (gradient or non-gradient based) methods, which are considerably sensible to initialization and may be stuck in local minima. As discussed in Section \ref{sec:previousWork}, in this work we model rigid slice-to-volume registration as a discrete labeling problem following the discretization strategy proposed by \cite{Zikic2010a}.

\subsection{Rigid Slice-to-Volume Registration as a Discrete Labeling Problem}

Rigid slice-to-volume registration, as well as many other problems in computer vision, can be formulated as a discrete labeling problem on a pairwise Markov Random Field (MRF) \cite{Wang2013}. Formally, a discrete pairwise MRF is an undirected graph $\mathcal{G} = \langle \mathcal{V}, \mathcal{E} \rangle$, where each node $v_i \in \mathcal{V}, i=1 ... |\mathcal{V}|$ represents a discrete variable. Any two variables $v_i, v_j$ depend on each other if there is an edge $(v_i, v_j) \in \mathcal{E}$ linking the corresponding nodes. The range of values that can be assigned to a discrete variable is determined by the label space $L$. A discrete labeling problem on a pairwise MRF consists on assigning a label $l_i \in L$ to every $v_i \in \mathcal{V}$, such that the following energy is minimized:

\begin{equation}
	\mathcal{P}(\mathbf{x}; G, F) = \sum_{v_i \in V} g_i(l_i) + \sum_{(v_i,v_j) \in \mathcal{E}} f_{ij}(l_i, l_j),
	\label{eq:mrfEnergy}
\end{equation}

\noindent where $\mathbf{x}=\{l_1, ..., l_n\}$ is a labeling assigning a label $l_i$ to every $v_i \in \mathcal{V}$, $G = \{g_i(\cdot)\}$ are the unary potentials associated to $v_i \in \mathcal{V}$ and $F = \{f_{ij}(\cdot, \cdot)\}$ are the pairwise potentials associated to edges $(v_i,v_j) \in \mathcal{E}$. These functions return scalar values when labels $l_i$ are assigned to variables $v_i$. Since we pose the optimization as a minimization problem, potentials must associate lower values to labelings representing good solutions, and higher values otherwise. 

In the formulation presented in Eq~\ref{eq:sliceToVolume}, one would like to explore the space of parameters $\pi$ and chose the values giving the best matching. Since we are modeling it as a discrete problem, we must adopt a discretization strategy for the (naturally) continuous space of rigid transformations $\pi$. In \cite{Zikic2010a}, authors proved that it is possible to estimate linear (an particularly, rigid body) transformations by solving a discrete and approximated version of this formulation. Following their proposal, we model rigid slice-to-volume registration through a graph $\mathcal{G} = \langle \mathcal{V}, \mathcal{E} \rangle$, associating every parameter of the rigid transformation $\pi = (r_x, r_y, r_z, t_x, t_y, t_z)$ to a variable $v_i \in \mathcal{V}$, giving a total of 6 variables (nodes in the graph). $\mathcal{G}$ is a fully connected pairwise graph where $\mathcal{E}=\{(v_i, v_j)\},  \forall i \neq j$, meaning that all variables (parameters) depend on each other. Note that this pairwise model is clearly an approximation, since the real dependency between the parameters is not pairwise but high-order. However, as stated in \cite{Zikic2010a}, similar approximations have shown to be good enough to estimate linear transformations, while making the problem tractable.

In our discrete strategy, every parameter $v_i$ is updated through a discrete variation $d_{l_i}$ associated to the label $l_i$. Given an initial transformation $\pi_0 = (r^0_x, r^0_y, r^0_z, t^0_x, t^0_y, t^0_z)$, we explore the space of solutions by sampling discrete variations of $\pi_0$, and choosing the one that generates the slice $\pi[J]$ best matching image $I$. Therefore, for a maximum size $\omega_i$ and a quantization factor $\kappa_i$, we consider the following variations to the initial estimate of $v_i$: $\{0, \pm \frac{\omega_i}{\kappa_i}, \pm  \frac{2\omega_i}{\kappa_i}, \pm \frac{3\omega_i}{\kappa_i}, ..., \pm \frac{\kappa_i \omega_i}{\kappa_i} \}$. The total number of labels results $|L| = 2 \kappa + 1$. Note that 0 is always included since we give the possibility of keeping the current parameter estimate. For example, in case that $v_0$ corresponds to $r_x$, $\omega_0=0.2$ and $\kappa_0=2$, the label space of $v_0$ will correspond to $\{r^0_x, r^0_x \pm 0.1, r^0_x \pm 0.2\}$.

Ideally, we would like to explore the complete search space around $\pi_0$ given by all possible combinations of labels. Since we have an exponential number of potential solutions, we adopt a pairwise approximation where only variations for pairs of variables are considered. This variations are encoded in the pairwise terms of the energy defined in Eq~\ref{eq:mrfEnergy} as $f_{ij}(l_i, l_j) = \mathcal{M}(I, \pi_{l_i, l_j}[J])$. Here $\pi_{l_i, l_j}$ denotes the updated version of $\pi_0$, where only $v_i$ and $v_j$ were modified according to the labels $l_i$ and $l_j$, while the rest of the parameters remained fixed. Unary potentials $g_i$ are not considered since we are only interested in the interaction between variables. Therefore, the discrete version of the optimization problem introduced in Eq~\ref{eq:sliceToVolume} becomes:

\begin{equation}
	\hat{\mathbf{x}} = \argmin_{\mathbf{x}} \mathcal{P}(\mathbf{x}; F) = \argmin_{\mathbf{x}} \sum_{(v_i,v_j) \in \mathcal{E}} \mathcal{M}(I, \pi_{l_i, l_j}[J]),
	\label{eq:sliceToVolumeMRF}
\end{equation}
where the optimal labeling $\hat{\mathbf{x}}$ represents the final rigid transformation $\hat{\pi}$ used to extract the solution slice $\hat{\pi}[J]$. 

\subsection{Discrete Optimization}
We solve the discrete multi-labeling problem from Eq~\ref{eq:sliceToVolumeMRF} using FastPD. FastPD is a discrete optimization algorithm based on principles from linear programming and primal dual strategies, which at the same time generalizes the well known $\alpha$-expansion \cite{Komodakis2008}. One of the main advantages of FastPD is its modularity/scalability, since it deals with a much wider class of problems than $\alpha$-expansion, being an order of magnitude faster while providing the same optimality guarantees when performing metric labeling \cite{Komodakis2007a}. Our problem does not fulfill the conditions to be considered a metric labeling problem (we refer the reader to \cite{Boykov2001} for a complete discussion about metric labeling); however, FastPD has shown promising results for similar formulations \cite{Zikic2010a}.

FastPD solves a series of max-flow min-cut \cite{Boykov2004} problems on a graph. In that sense, it is similar to $\alpha$-expansion which also performs discrete inference on multi-label problems by solving successive binary max-flow min-cut problems. The main difference between these approaches is the construction of the graph where max-flow min-cut algorithm is applied. $\alpha$-expansion constructs the binary problem by restricting the label space, so that the only options for a given variable are to remain in its current assignment, or to take a label $\alpha$ (which varies in every iteration). Instead, FastPD constructs these binary problems by performing a Linear Programming Relaxation (LPR) of the integer program that represents the discrete MRF formulation. 
\subsection{Incremental Approach}
Discrete approximations of continuous spaces usually suffer from low accuracy (since it is bounded by the quality of the discretization). Thus, we adopt an incremental approach to explore the space of solutions in a finer way. The idea is to successively solve the problem from Eq~\ref{eq:sliceToVolumeMRF}, using the solution from time $t$ as initialization for time $t+1$, keeping a fixed number of labels but decreasing the maximum sizes $\omega_i$ in a factor $\alpha_i$. Moreover, we also adopt a pyramidal approach, where we generate a Gaussian pyramid for both input images $I$ and $J$, and we run the complete incremental approach on every level of the pyramid. In that way, we increase the capture range of our method.
\begin{figure}[t]
  \centering
  \includegraphics[width=\textwidth]{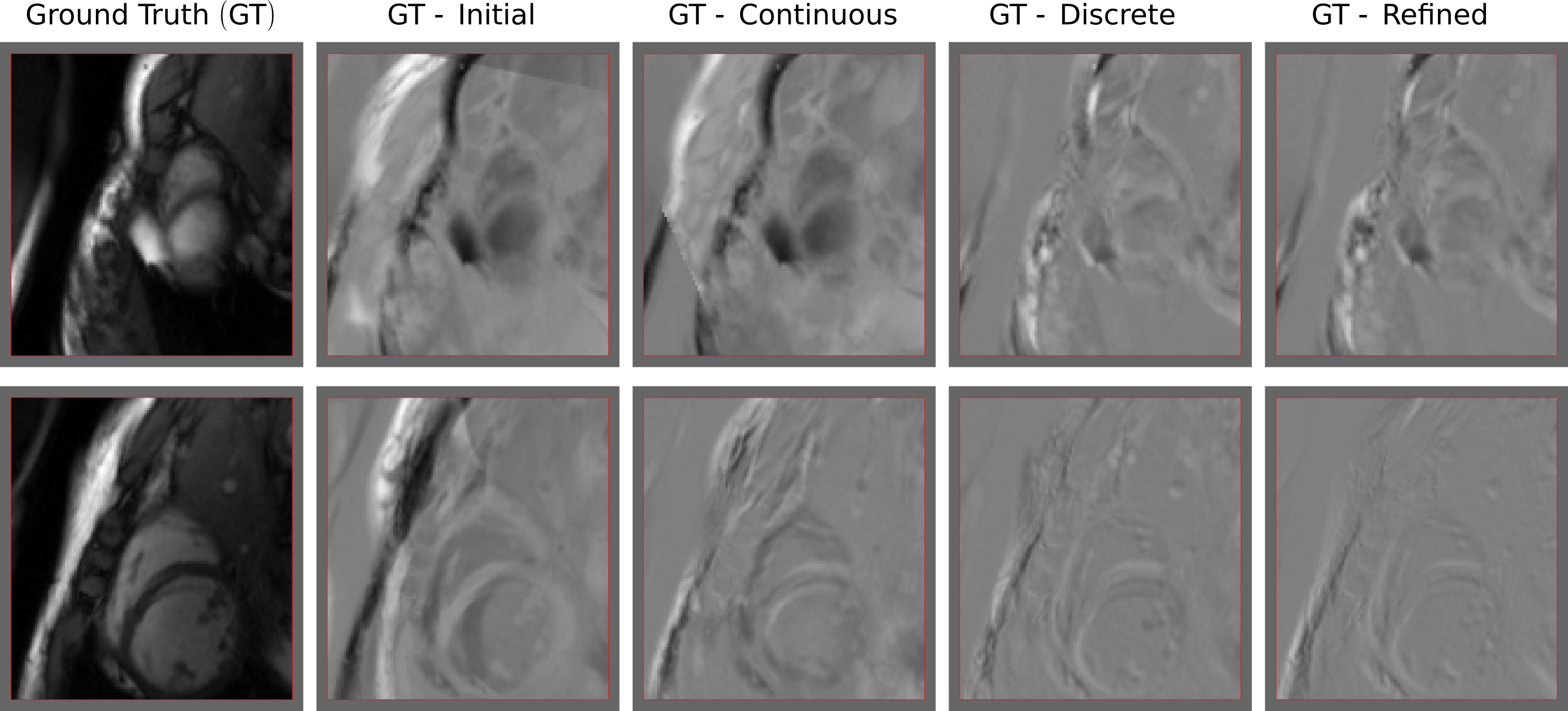}
  \caption{Visual results for two slices of the individual tests. The first column corresponds to the input 2D slice. The second column shows the difference between the input 2D slice and the initial slice. The other columns show the difference between the input and the one resulting slices applying simplex, discrete and refined approaches. Grey values indicate no difference, while white and black value indicate inconsistencies. As it can be observed, the solution given by the refined approach is outperforming the others.}
  \label{fig:visualResults}
\end{figure}
\subsection{Simplex Refinement Step}
Let us advance one of the conclusions of this work, so that we can motivate the last step of our approach. In Section \ref{sec:resultsIndividualTests}, we compare the performance of our discrete approach with a continuous method based on simplex \cite{Nelder1965} algorithm. As we will see, when the initialization $\pi_0$ is good enough, both continuous and discrete approaches perform well. In fact, in some cases, simplex is delivering more accurate solutions than discrete. However, as we move away from the initialization, discrete optimization gives more and more significant improvements, thanks to its wider capture range. In order to improve the accuracy of our proposal, and inspired by similar conclusions discussed by \cite{Lempitsky2008} in the context of vector flow estimation, we refine the results obtained with our approach by optimizing Eq \ref{eq:sliceToVolume} through simplex, using the discrete solution as initialization. 
\section{Experiments and Results}
\label{sec:results}
In this section, we present the results obtained using the proposed method (considering also the refined version), and we compare them with a state-of-the-art approach based on continuous optimization trough downhill simplex \cite{Nelder1965} (also known as Nelder-Mead or amoeba method). Simplex is one of the most popular optimization algorithms used to deal with rigid slice-to-volume registration (some examples are \cite{Xu2014b, Kim1999, Birkfellner2007, Park2004}). It is a continuous and derivative-free method, which relies on the notion of simplex (which is a special polytope of $n+1$ vertices living in a $n$-dimensional space) to explore the space of solutions in a systematic way. We used a dataset composed of MRI images of a beating heart. Given an initial sequence of 3D images $M_i, i=0 .. 19$ of a beating heart (with a resolution of $192 \times 192 \times 11$ voxels and a voxel size of $1.25 \times 1.25 \times 8 mm$), we generated slices which were used for two different experiments. 
\subsection{Implementation Details and Parameters Setting}
We implemented the three versions of the algorithms discussed in this paper (simplex, discrete and refined) mainly using Python and ITK for image manipulation \footnote{The source code can be downloaded from \\ \texttt{
https://gitlab.com/franco.stramana1/slice-to-volume}}. For simplex optimization we used the method implemented in \textit{scipy.optimize} package, while discrete optimization was performed using a Python wrapped version of the standard C++ implementation of FastPD. In all the experiments, we used a pyramidal approach with 4 Gaussian levels (3D images where not downsampled in z axis because of the low resolution of the images in this direction). The matching criterion adopted in all the experiments was the sum of squared differences, since pixel intensities are equivalent in both 2D and 3D images. The matching criterion $\mathcal{M}$ based on SSD is simply defined as:
\begin{equation}
	\mathcal{M}(I_1, I_2) = \sum_{i \in \Omega} (I_1(i) - I_2(i))^2,
	\label{eq:matchingCriterion}
\end{equation}

For the discrete case, at every pyramid level we decreased the maximum label size for both, rotation ($\omega_{rot} = [$0.02, 0.015, 0.0125, 0.01$]$rad) and translation ($\omega_{trans} = [$7, 6.5, 6, 5$]$mm) parameters. Starting from these maximum sizes, we solved Eq~\ref{eq:sliceToVolumeMRF} running FastPD several times per level until no improvement is produced or a maximum number of iterations is achieved ([200, 100, 150, 600]), using different label space decreasing factors at every pyramid level ($\alpha$=[0.08, 0.07, 0.05, 0.03]). The total number of labels was fixed to 5 ($\kappa = 2$) for all the experiments. For the continuous case (where Eq~\ref{eq:sliceToVolume} was optimized using simplex), we used again a 4-levels pyramidal approach, with simplex running until convergence in every level. Finally, for the refined experiment, we just ran the simplex experiment initialized with the solution estimated with the discrete method. For every registration case, continuous approach took around 30secs while the discrete version took 9mins, running on a laptop with an Intel i7-4720HQ and 16GB of RAM.
\label{sec:resultsIndividualTests}
\begin{figure}[t]
  \centering
  \includegraphics[width=\textwidth]{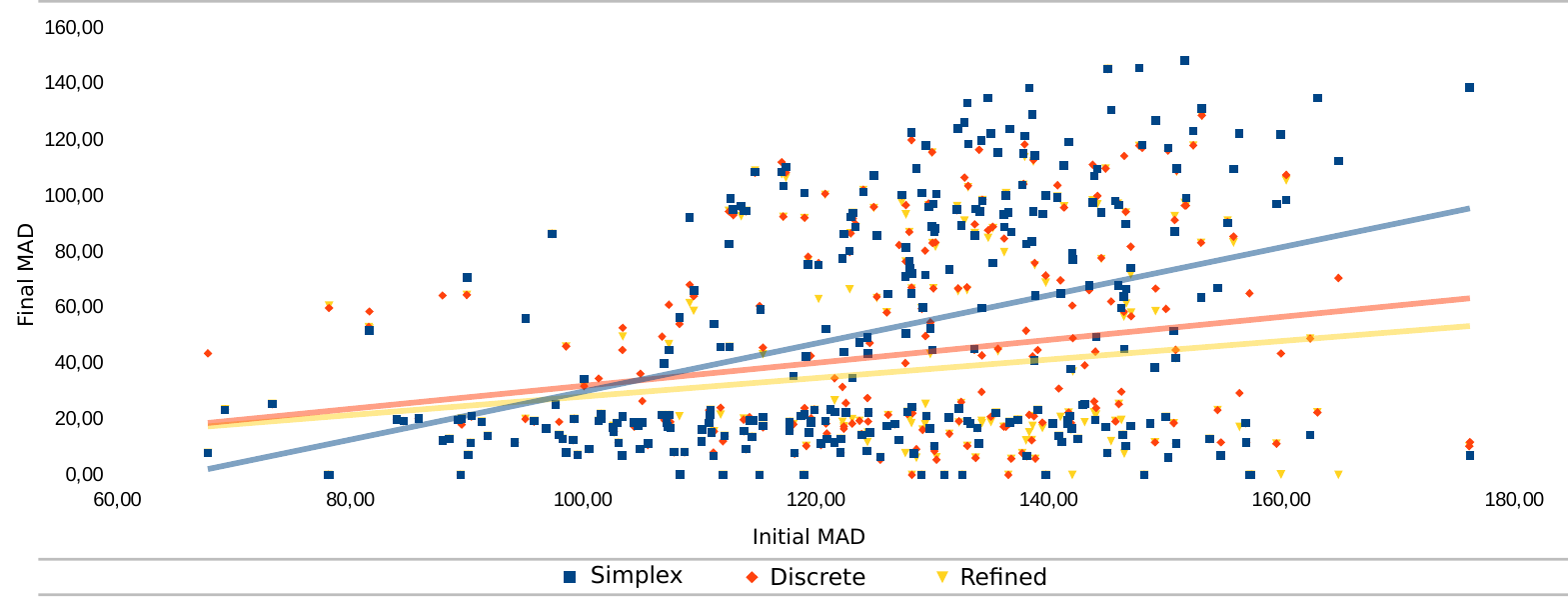}
  \caption{Individual tests where 100 2D slices (extracted at locations specified using random rigid transformations) are considered as independent registration cases. Every point form the scatter plot represents the mean of absolute differences (MAD) between the input 2D image and the slice extracted at the initial position (X axis) vs the estimated position (Y axis). We also include a linear trend estimation (fitted using least squares method) to compare the robustness of the method to bad initializations.}
  \label{fig:individualSAD}
\end{figure}
\subsection{Experiments}
We performed two different type of experiments, considering individual registration cases as well as image series. For validation, we measured three different indicators: the distance in terms of translation and rotation parameters between the estimated and ground truth transformations, together with the mean of absolute differences (MAD) between the input 2D image and the slice specified by the estimated rigid transformation.\\
\begin{figure}[t]
  \centering
  \includegraphics[width=\textwidth]{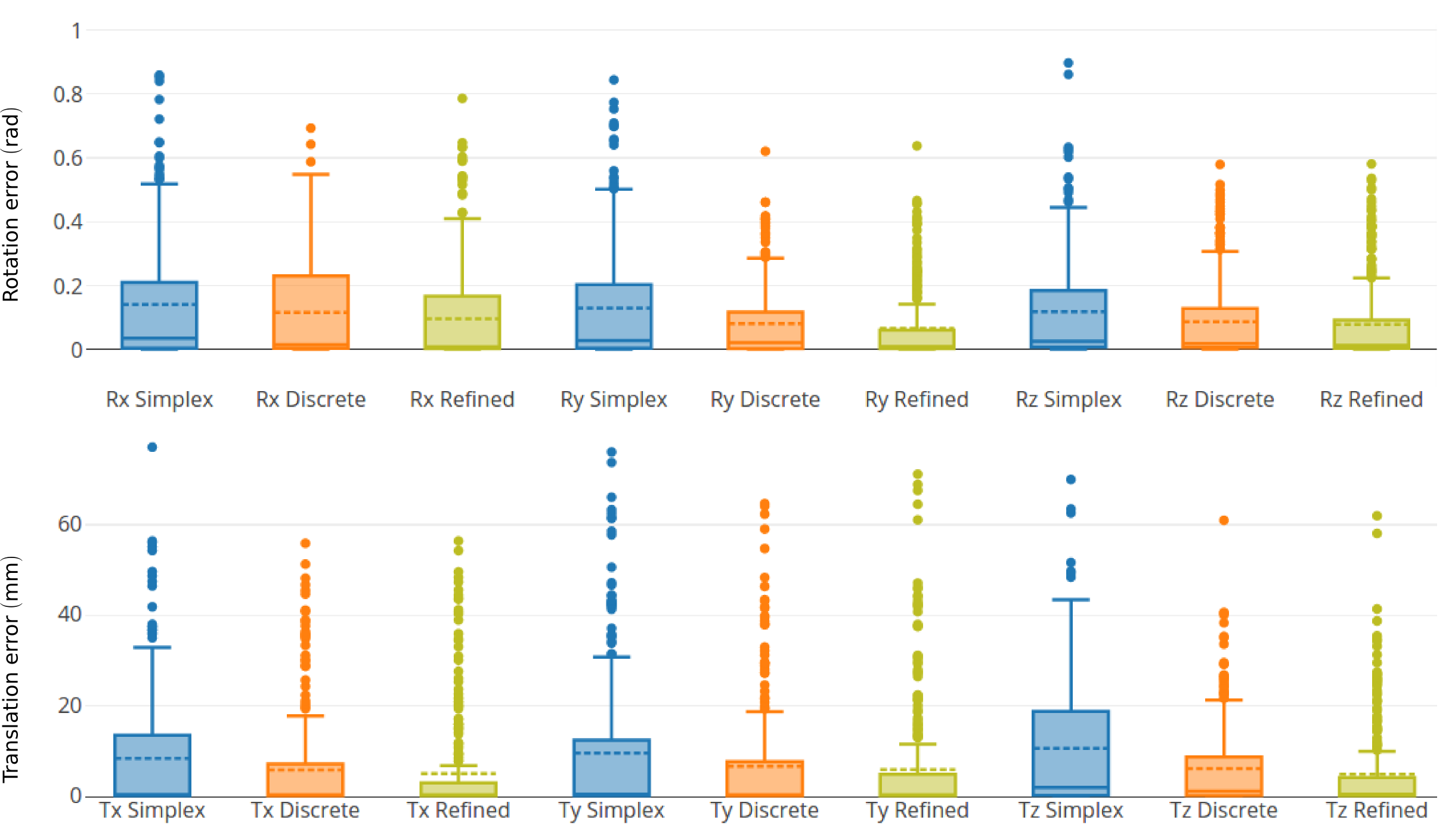}
  \caption{Boxplot corresponding to the estimated error (in terms of rotation and translation parameters) for the 300 individual tests. As it can be observed, discrete and refined approaches are reducing both the mean error (shown as a dotted line in every box) and the dispersion.}
  \label{fig:individualBoxplot}
\end{figure}

\begin{table}[t]
\centering
\label{tab:individualTests}

\caption{Average error estimated in terms of rotation (expressed in radians), translation (expressed in millimeters) and MAD for the three alternative approaches discussed in this paper. As it can be observed, the discrete and refined methods outperform the standard continuous approach optimized through simplex.}

\begin{tabular*}{\columnwidth}{@{\extracolsep{\stretch{1}}}*{8}{c}@{}}
\hline
\textbf{Method}   & $R_x$         & $R_y$         & $R_z$         & $T_x$         & $T_y$         & $T_z$         & \textit{MAD}   \\ \hline
\textit{Simplex}  & 0,14          & 0,13          & 0,12          & 8,46          & 9,62          & 10,69         & 51,88          \\ \hline
\textit{Discrete} & 0,12          & 0,08          & 0,09          & 5,87          & 6,72          & 6,18          & 42,36          \\ \hline
\textit{Refined}  & \textbf{0,10} & \textbf{0,07} & \textbf{0,08} & \textbf{5,09} & \textbf{5,96} & \textbf{4,92} & \textbf{36,45} \\ \hline
\end{tabular*}

\end{table}

\noindent \textbf{Individual tests.} The first set of experiments measures the accuracy of the three approaches using individual tests, where 100 random slices extracted from the 20 volumes are considered as single images (independently of the series), and registered to the first volume $M_0$. We run the same experiment for every slice using three different initializations (resulting in 300 registration cases), where ground truth parameters were randomly perturbed in three different ranges ([5, 12), [12, 18), [18, 25) millimeters for translation and [0.1 , 0.2), [0.2, 0.3), [0.3, 0.4) radians for rotation parameters) to guarantee that both, good and bad initializations, are considered for every slice. Quantitative results are reported in Figures \ref{fig:individualSAD}, \ref{fig:individualBoxplot} and summarized in Table \ref{tab:individualTests}. Visual results for qualitative evaluation are reported in Figure \ref{fig:visualResults}.

Results in the scatter plot from Figure \ref{fig:individualSAD} indicate that, as we go farther away from the initialization (in this case, it is quantified by the MAD between the input 2D image and the slice corresponding to the initialization), discrete and refined methods tend to be more robust. This robustness is clearly reflected by the slope of the trend lines: the refined method presents the trend line with the lower slope, meaning that even for bad initializations it converges to better solutions. The boxplot from Figure \ref{fig:individualBoxplot} and the numerical results from Table \ref{tab:individualTests} confirm that discrete and refined methods perform better not only in terms of MAD, but also with respect to the distance between the rotation/translation estimated and ground truth parameters.\\

\begin{figure}[t]
  \centering
  \includegraphics[width=\textwidth]{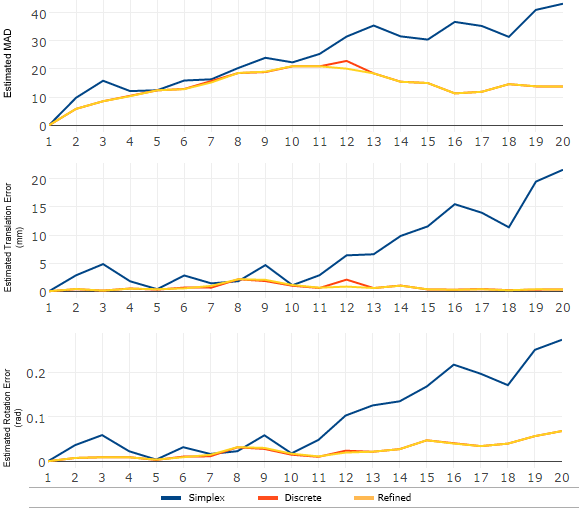}
  \caption{Results for the temporal series experiment. In this case, the transformation estimated for the slice $i$ was used as initialization for the next slice of the series. We reported results in terms of MAD and rotation/translation error for the estimated transformations using the three approaches.}
  \label{fig:temporalSeries}
\end{figure}

\noindent \textbf{Temporal series test.} The idea behind the second experiment is to simulate an image guided surgery (IGS) scenario, where a fixed pre-operative volume must be fused with consecutive intra-operative 2D images suffering deformations (in this case, due to heart beating). Given the temporal sequence of 20 volumetric MRI images $M_i, i=0 .. 19$, we generated a sequence of 20 2D slices to validate our method. It was extracted as in \cite{Ferrante2013}: starting from a random initial translation $T_0=(T_{x_0}, T_{y_0}, T_{z_0})$ and rotation $R_0=(R_{x_0}, R_{y_0}, R_{z_0})$, we extracted a 2D slice $I_0$ from the initial volume $M_0$. Gaussian noise was added to every parameter in order to generate the position used to extract the next slice from the next volume. We used $\sigma_r=3^\circ$ for the rotation and $\sigma_t=5\mathit{mm}$ for the translation parameters. All the slices were registered to the first volume $M_0$. The solution of the registration problem for slice $I_i$ was used as initialization for the slice $I_{i+1}$. The first experiment was initialized randomly perturbing its ground truth transformation with the same noise parameters. As it can be observed in Figure \ref{fig:temporalSeries}, discrete and refined approaches manage to keep a good estimation error while simplex can not. Note that different strategies could be used in real scenarios to obtain good initializations for the first slice. For example, in IGS, physicians could start from a plane showing always the same anatomical structure, or identify landmark correspondences in the first slice and the 3D image useful to estimate an initial transformation.

\section{Discussion, Conclusions and Future Works}
In this paper we presented a strategy to solve rigid slice-to-volume registration as a discrete graph labeling problem, following the discretization strategy introduced by \cite{Zikic2010a}. We validated our proposal using a MRI dataset of a beating heart, where arbitrary 2D slices are fused with a 3D volume. The experimental results showed that our discrete approach produces more accurate and robust estimates for the rigid transformations than a continuous method based on simplex. Moreover, they also reflected that results obtained using such a method can be further refined using a continuous approach like simplex, leading to even more accurate estimations. This is coherent with the conclusions presented by \cite{Lempitsky2008} for the case of optical flow estimation.

An interesting discussion about the limitations of our approach, emerges when we observe the results obtained in previous work by \cite{Ferrante2013, Ferrante2015, Ferrante2015a} using similar images. In these works, both rigid and local deformable parameters are estimated in a one-shot discrete optimization procedure, delivering results which are considerably better than ours, even for the refined approach. Since we are dealing with 2D images which are deformed with respect to the static 3D volume (due to heart beating), estimating both rigid and deformable parameters at the same time seems to be the correct solution since there is a clear mutual dependence between them. However, if we look at the results corresponding to the first slices of the temporal series in Figure \ref{fig:temporalSeries} (where there is almost no deformation, and even null deformation for the 1st slice), we can see that the quality of the solution is significantly better than in the other cases. In fact, the error is almost 0. It suggests that when the 2D image is not deformed with respect to the input volume, our method is enough to capture slice-to-volume mapping. This limitation is somehow inherent to the model we are using: rigid transformations can not deal with local deformations. To improve the results in these cases, we plan to extend our approach to linear transformations where also anisotropic scaling and shearing can be considered. Following the strategy by \cite{Zikic2010a}, it will result straightforward.

Finally, a future line of research has to do with applying discrete rigid (or linear) slice-to-volume registration to other problems. As discussed in Section \ref{sec:motivation} motion correction for volume reconstruction is another problem requiring mapping slice-to-volume. It would be interesting to explore how our approaches performs in this case. 

\bibliographystyle{splncs}
\bibliography{library}

\end{document}